\documentclass[letterpaper]{article}
\usepackage{subfigure}
\usepackage{multirow}
\usepackage{graphicx}
\usepackage{amssymb}
\usepackage{amsmath}
\usepackage{aaai}
\usepackage{times}
\usepackage{helvet}
\usepackage{courier}
\begin{document}
%
\title{Coupled Item-based Matrix Factorization}
\author{Fangfang Li, Guandong Xu, Longbing Cao\\
Advanced Analytics Institute\\
University of Technology, Sydney, Australia\\
}
\maketitle
\begin{abstract}
The essence of the challenges \textit{cold start} and \textit{sparsity} in Recommender Systems (RS) is that the extant techniques, such as Collaborative Filtering (CF) and Matrix Factorization (MF), mainly rely on the user-item rating matrix, which sometimes is not informative enough for predicting recommendations. To solve these challenges, the objective item attributes are incorporated as complementary information. However, most of the existing methods for inferring the relationships between items assume that the attributes are ``independently and identically distributed (iid)", which does not always hold in reality. In fact, the attributes are more or less coupled with each other by some implicit relationships. Therefore, in this paper we propose an attribute-based coupled similarity measure to capture the implicit relationships between items. We then integrate the implicit item coupling into MF to form the Coupled Item-based Matrix Factorization (CIMF) model. Experimental results on two open data sets demonstrate that CIMF outperforms the benchmark methods.
\end{abstract}

\section{Introduction}
Recommender Systems (RS) are proposed to help users tackle information overload by suggesting potentially interesting items to users \cite{melville_RS}. The main challenges in RS now are \textit{cold start} and \textit{sparsity} problems. The essence behind these challenges is that traditional recommendation techniques such as Collaborative Filtering (CF) or Matrix Factorization (MF) normally rely on the user-item rating matrix only, which sometimes is not informative enough for making recommendations.

To solve these challenges, researchers attempt to leverage complementary information, such as social friendships, in RS. For instance, many social recommender systems \cite{DBLP:MaYLK08a_socialRS_PMF} \cite{DBLP:MaKL09_SocialTrust} \cite{DBLP:MaZLLK11_SocialRegularization} \cite{DBLP:YangSL12_circleRS} \cite{DBLP:MaoYe_Exploring_socia_influence} have been proposed utilizing social friendships. To some extent, the social friendships in RS have been well studied but most web sites do not have social mechanisms, therefore, several researchers are currently trying to incorporate users' or items' attributes to improve the performance of recommendation. For example, some have tried to estimate the latent factors through considering the attributes \cite{Rendle10_factorizationMachine} \cite{GantnerDFRS10_attributeMapping} \cite{MenonE10_logLinearlatent} \cite{AgarwalC09_regression-mf} or topic information of users and items \cite{AgarwalC10_flda} \cite{WangB11_collaborative_topic_modeling} \cite{HU_implicitFeedback_CF} for latent factor models. Nevertheless, most of the existing methods assume that the attributes are independent. In reality, however this assumption does not always hold and there exist complex coupling relations between instances and attributes. For example, in Table \ref{tab:toy-example}, the ``Director", ``Actor", and ``Genre" attributes in movies are often coupled together and influence each other. Therefore, in this paper, we deeply analyze the couplings between items to capture their implicit relationships, based on the attribute information.

\begin{table}[htbp]
  \centering
  \caption{A Toy Example}
    \begin{tabular}{|c| c c c c|}
    \cline{1-5}
     Director & Scorsese & Coppola & Hitchcock & Hitchcock\\
    Actor & De Niro & De Niro & Stewart & Grant\\
     Genre & Crime & Crime & Thriller & Thriller\\
    \hline
    \hline
     & God & Good & Vertigo & N by\\
     & Father & Fellas & & NW\\
    \hline
    $u_1$  & 1 & 3 & 5 & 4 \\
    $u_2$ & 4 & 2 & 1 & 5\\
    $u_3$ & - & 2 & - & 4\\
    \hline
    \end{tabular}
  \label{tab:toy-example}
\end{table}

We know that the implicit relationships can be aggregated from the similarity of attribute values for all the attributes. From the perspective of ``iid" assumption, different attribute values are independent, and one attribute value will not be influenced by others. However, if we disclose the ``iid" assumption, we will observe that one attribute value will also be dependent on other values of the same attribute. Specifically, two attribute values are similar if they present the analogous frequency distribution on one attribute, which leads to another so-called intra-coupled similarity within an attribute. For example, two directors ``Scorsese" and ``Coppola" are considered similar because they appear with the same frequency. On the other hand, the similarity of two attributes values is dependent on other attribute values from different attributes, for example, two directors' relationship is dependent on ``Actor" and ``Genre" attributes over all the movies. This dependent relation is called the inter-coupled similarity between attributes. We believe that the intra and inter-coupled similarities disclosing the ``iid" assumption should simultaneously contribute to analysing the relationships between items, namely item coupling.

Rating preferences have been well studied but the relationships between items, especially implicit relations, are still far away from being successfully incorporated into the MF model. Therefore, in this paper, we propose a Coupled Item-based MF (CIMF) model incorporating implicit item couplings through a learning algorithm via regularization on implicit and explicit information. After accommodating the implicit relationships between items and users' preferences on items into a unified learning model, we can predict more satisfactory recommendations, even for new users/items or when the rating matrix is very sparse. The motivation for incorporating such couplings into RS is to solve the \textit{cold start} and \textit{sparsity} challenges in RS by leveraging the items' implicit objective couplings and users' subjective rating preferences on items. When we have ample rating data, the user-item rating matrix is mainly applied for recommendation. However, for new users or items of RS or for a very sparse rating matrix, implicit item coupling would be mainly exploited for making recommendations.

The contributions of the paper are as follows:
\begin{itemize}
\item We propose a NonIID-based method to capture the implicit relationships between items, namely item coupling, based on the their objective attribute information.
\item We propose the CIMF model which integrates the item coupling and users' subjective rating preferences into a matrix factorization learning model.
\item We conduct experiments to evaluate the effectiveness of our proposed CIMF model.
\end{itemize}

The rest of the paper is organized as follows. Section 2 presents the related work. In Section 3, we formally state the recommendation and couplings problems. Section 4 first analyzes the couplings in RS, after which it details the coupled Item-based MF model integrating the couplings together. Experimental results and the analysis are presented in Section 5. The paper is concluded in the final section.

\section{Related Work}

The approaches related to our work in recommender systems include collaborative filtering and content-based techniques.

\subsection{Collaborative Filtering}
Collaborative filtering (CF)\cite{SuK09_survey_CF} \cite{Sarwar:2001:ICF} \cite{Deshpande:2004:ITN_recommendation} is one of the most successful approaches, taking advantage of user rating history to predict users' interests. User-based CF and item-based CF are mainly involved in the CF method. The basic idea of user-based CF is to recommend interesting items to the active user according to the interests of the other users with whom they have close relationships. Similarly, item-based CF tries to recommend to the active user potentially interesting items which have close similarities with the historical items that the active user likes. As one of the most accurate single models for collaborative filtering, matrix factorization (MF)\cite{DBLP:Koren08_Factorization} \cite{DBLP:KorenBV09_MF} is a latent factor model which is generally effective at estimating the overall structure that relates simultaneously to most items. The MF approach tries to decompose the rating matrix to the user intent matrix and the item intent matrix. Then, the estimated rating is predicted by the multiplication of the two decomposed intent matrices.

Although there had been wide adoption of this approach in many real applications, e.g., Amazon, the effect of CF is sharply weakened in the case of new users or items and for a very sparse rating matrix. This is partly because when the rating matrix is very sparse, for new users or items, it is extremely difficult to determine the relationships between users or items. This limitation partly motivates us to consider implicit item coupling to enhance the effectiveness of recommendations.

\subsection{Content-based Methods}
Content-based techniques are another successful method by which to recommend relevant items to users by matching the users' personal interests to descriptive item information \cite{TKDE.2005.99_nextRS} \cite{Mooney:2000:CBR} \cite{Pazzani07cbr}. Generally, content-based methods are able to cope with the sparsity problem, however, they often assume an item's attributes are ``iid" which does not always hold in reality. Actually, several research outcomes \cite{DBLP:CaoOY12_CBA} \cite{canWang_coupledClustering} \cite{CanWang_IJCAI} \cite{WangSC13} have been proposed to handle these challenging issues. To the best of our knowledge, in relation to RS, there is only one paper \cite{YonghongYu_coupledItemsRecommedation} which applies a coupled clustering method to group the items, then exploits CF to make recommendations. But from the perspective of RS, this paper does not fundamentally disclose the ``iid" assumption for items. This motivates us to analyze the intrinsic relationships from different levels to unfold the assumption.

\section{Problem Statement}
A large number of user and item sets with specific attributes can be organized by a triple $S=<U,S_O,f>$, where $U=\{u_1, u_2,..., u_n\}$ is a nonempty finite set of users, $S_O = <O,A,V,g>$ describes the items' attribute space. Among $S_O$, $O=\{o_1, o_2,..., o_m\}$ is a nonempty finite set of items, $A = \{A_1,...,A_M\}$ is a finite set of attributes for items; $V=\cup_{j=1}^J{V_j}{}$ is a set of all attribute values for items, in which $V_j$ is the set of attribute values of attribute $A_j (1\le j \le J)$, $V_{ij}$ is the attribute value of attribute $A_j$ for item $o_i$, and $g=\land_{j=1}^Mg_j (g_j:U \to V_j)$ is an mapping function set which describes the relationships between attribute values and items. In the triple $S=<U,S_O,f>$, $f(u_i,o_j) = r_{ij}$ expresses the subjective rating preference on item $o_j$ for user $u_i$. Through the mapping function $f$, user rating preferences on items are then converted into a user-item matrix $R$, with $n$ rows and $m$ columns. Each element $r_{ij}$ of $R$ represents the rating given by user $u_i$ on item $o_j$. For instance, Table \ref{tab:toy-example} consists of three users $U=\{u_1,u_2,u_3\}$ and four items $O=\{God Father, Good Fellas, Vertigo, N by NW\}$. The items have attributes $A=\{Director, Actor, Genre\}$, and attribute values $V_3=\{Crime,Thriller\}$. The mapping functions are $g_3(Vertigo)= Thriller$ and $f(u_2,Vertigo) = 1$.

\begin{figure}
\centerline{\includegraphics[scale=0.7]{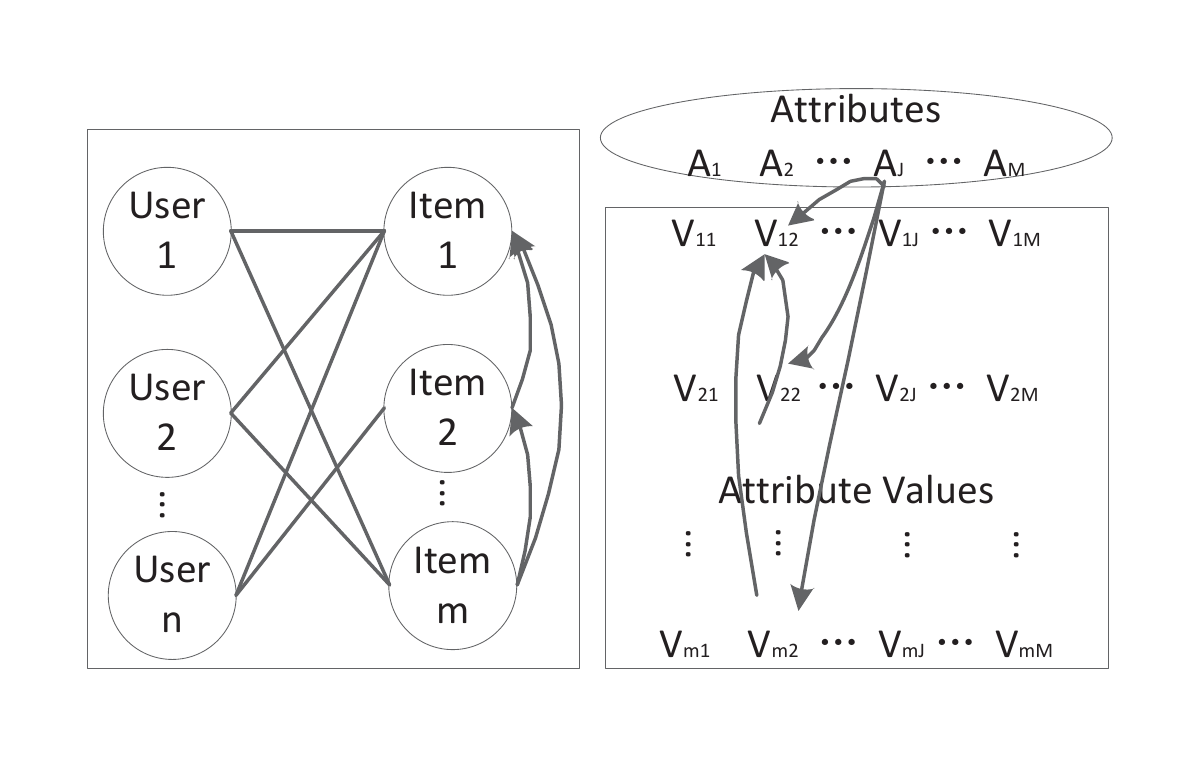}}
\caption{Item Couplings in Recommender Systems}
\label{fig_Coupling}
\end{figure}

As mentioned, the extant similarity methods for computing the implicit relationships within items assume that the attributes are independent of each other. However, all the attributes should be coupled together and further influence each other. The couplings between items are illustrated in Fig.\ref{fig_Coupling}, which shows that within an attribute $A_j$, there is dependence relation between values $V_{lj}$ and $V_{mj}$ $(l \not=  m)$, while a value $V_{li}$ of an attribute $A_i$ is further influenced by the values of other attributes $A_j$ $(j \not= i)$. For example, attributes $A_1$, $A_3$, ... to $A_J$ all more or less
influence the values of $V_{12}$ to $V_{n2}$ of attribute $A_2$.

In order to capture the implicit item coupling and disclose the ``iid" assumption, we first introduce several basic concepts as follows.

Given an attribute space $S_O=<O,A,V,g>$, the \textbf{Intra-coupled Attribute Value Similarity (IaAVS)} between values $x$ and $y$ of attribute $A_j$ for items is defined as:
\begin{equation}
\delta^{Ia}_j(x,y)=\frac{|g_j(x)|.|g_j(y)|}{|g_j(x)|+|g_j(y)|+|g_j(x)|.|g_j(y)|}
\label{eqn-intra-coupling}
\end{equation}
where $g_j(x) = \{o_i|V_{ij} = x, 1\le j \le M,1\le i\le n \}$ is the subset of $O$ with corresponding attribute $A_j$ having attribute value $x$, and $|g_j(x)|$ is the size of the subset.

The influence of attribute value $y$ of attribute $A_k$ for attribute value $x$ of attribute $A_j$ can be calculated by:
\begin{equation}
P_{j|k}(y|x) = \frac{|g_{j,k}(x,y)|}{|g_j(x)|}
\end{equation}
where $g_{j,k}(x,y) = \{o_i|(V_{ij} = x) \land (V_{ik} = y), 1\le j,k \le M, 1\le i\le n \}$

Given an attribute space $S_O=<O,A,V,g>$, the \textbf{Inter-coupled Relative Similarity (IRS)} between attribute values $x$ and $y$ of attribute $A_j$ based on another attribute $A_k$ is:
\begin{equation}
\delta_{j|k}(x,y) = \sum\limits_{w\in\cap} {min\{P_{k|j}({w}|x), P_{k|j}({w}|y)\}}
\end{equation}
where $w$ is an attribute value for attribute $A_k$ of items with attribute $A_j$ having both values $x$ and $y$.

Given an attribute space $S_O=<O,A,V,g>$, the \textbf{Inter-coupled Attribute Value Similarity (IeAVS)} between attribute values $x$ and $y$ of attribute $A_j$ for item set $O$ is:
\begin{equation}
\delta^{Ie}_j(x,y) = \sum\limits_{k = 1,k \ne j}^n {\gamma_k \delta_{j|k}(x,y)}
\end{equation}
where $\gamma_k$ is the weight parameter for attribute $A_k$, $\sum\limits_{k=1,k\neq j}^n {\gamma_k} = 1$, $\gamma_k \in [0,1]$, and $\delta_{j|k}(x,y)$ is inter-coupled relative similarity.

Based on IaAVS and IeAVS, the \textbf{Coupled Attribute Value Similarity (CAVS)} between attribute values $x$ and $y$ of attribute $A_j$ is defined as follows.
\begin{equation}
\delta^A_j(x, y) = \delta^{Ia}_j(x, y) * \delta^{Ie}_j(x, y)
\end{equation}

\section{Coupled Item-based MF Model}
\begin{figure}
\centerline{\includegraphics[scale=0.7]{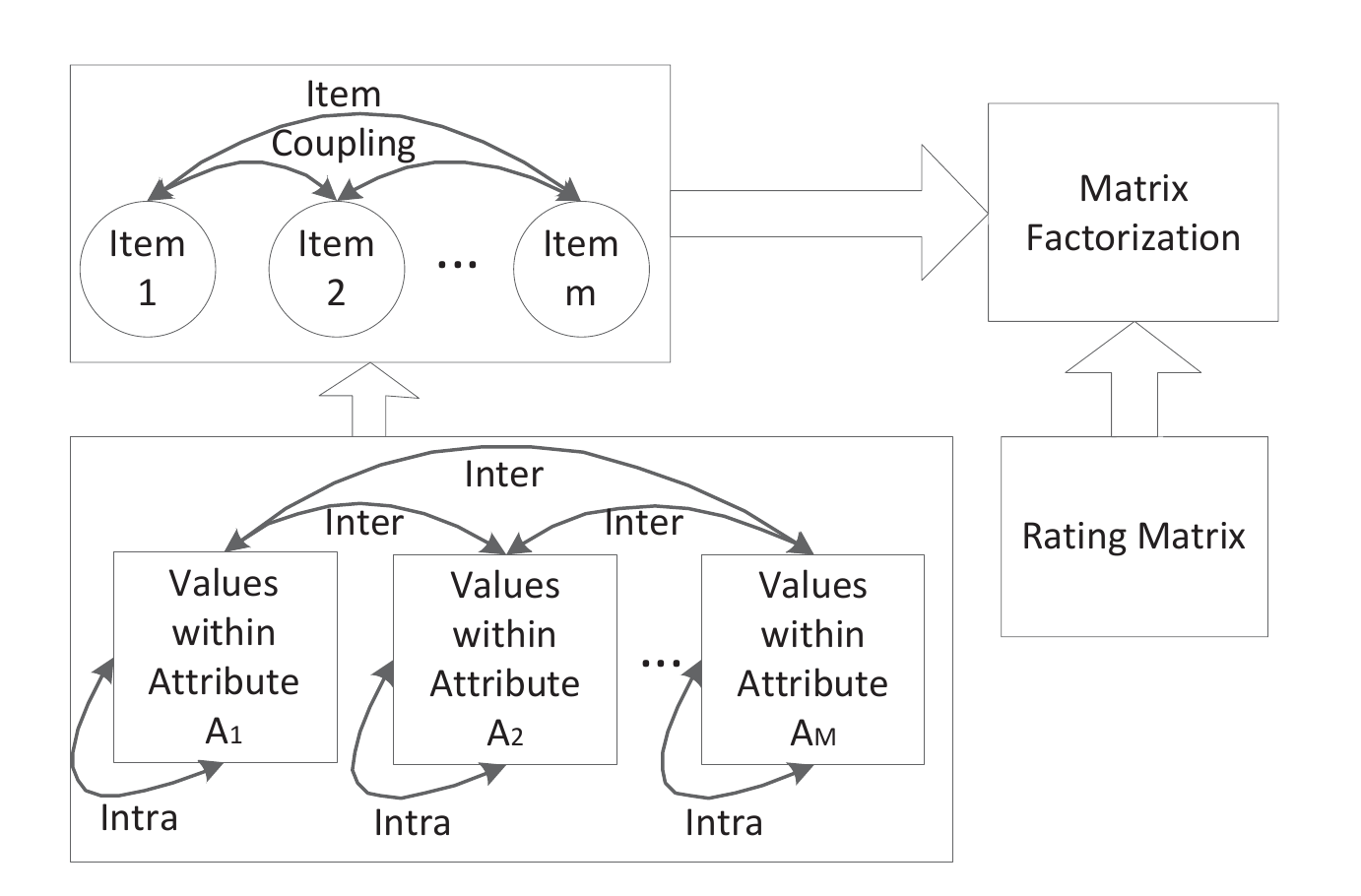}}
\caption{Coupled Item-based MF Model}
\label{fig_CoupledFramework-of-MF}
\end{figure}

In this section, we first introduce item coupling, then detail how to integrate it into the proposed CIMF model as in \ref{fig_CoupledFramework-of-MF}.

\subsection{Item Coupling}
As mentioned above, item coupling should reflect the implicit relationships between different items. For two items described by the attribute space $S_O=<O,A,V,g>$, the Coupled Item Similarity (\textit{CIS}) is defined to measure the similarity between items.

Formally, given item attribute space $S_O = <O,A,V,g>$, the \textbf{Coupled Item Similarity (\textit{CIS})} between two items $o_i$ and $o_j$ is defined as follows.
\begin{equation}
CIS(o_i,o_j) = \sum\limits_{k = 1}^{J} {\delta ^A_k(V_{ik}, V_{jk})}
\end{equation}
where $V_{ik}$ and $V_{jk}$ are the values of attribute $j$ for items $o_i$ and $o_j$, respectively; and $\delta ^A_k$ is Coupled Attribute Value Similarity.

From this definition, we clearly see that the intra-couplings between values within an attribute and inter-couplings between attributes are incorporated for measuring item coupling which partly helps to uncover the intrinsic
relations within items rather than considering them independently.


\subsection{Coupled Item-based MF Model}
Traditionally, the matrix of predicted ratings $\hat R\in{\mathbb{R}^{n\times m}}$, where $n$, $m$ respectively denote the number of users and the number of items, can be modeled as:

\begin{equation}
\hat R = {r_m} + P{Q^T}
\label{model-basicMF}
\end{equation}
with matrices $P\in{\mathbb{R}^{{n} \times d}}$ and $Q\in{\mathbb{R}^{{m} \times d}}$, where $d$ is the rank (or dimension of the latent space) with $d \le {n},{m}$, and $r_m \in \mathbb{R}$ is a global offset value. Through Eqn. \ref{model-basicMF}, the prediction task of matrix $\hat{R}$ is transformed to compute the mapping of users and items to factor matrices $P$ and $Q$. Once this mapping is completed, $\hat R$ can be easily reconstructed to predict the rating given by one user to an item by using Eqn. \ref{model-basicMF}. To avoid over-fitting, the regularization factors $\|P\|$ and $\|Q\|$ are added into the loss function to penalize over influence by observations.

On top of the traditional MF method, we propose a novel CIMF model which takes not only the rating preferences, but also item coupling into account. The learning procedure is constrained two-fold: the learned rating values should be as close as possible to the observed rating values, and the predicted item profiles should be similar to their neighbourhoods as well, which are derived from their implicit coupling information. More specifically, item coupling is incorporated by adding an additional regularization factor in the optimization step. Then, the computation of the mapping can be similarly optimized by minimizing the regularized squared error. The objective function is amended as Eqn. \ref{objctiveFuction-CMF}.

\begin{equation}\label{objctiveFuction-CMF}
\begin{split}
& L = \frac{1}{2}\mathop \sum \limits_{\left( {u,o_i} \right) \in K} {\left( {{R_{u,o_i}} - \hat R_{u,o_i}} \right)^2} + \frac{\lambda}{2}\left( {\|{Q_i\|^2} + \|{P_u\|^2}} \right)\\
&+\frac{\alpha }{2}\sum\limits_{all(o_i)} {{{\left\| {{Q_i} - \sum\limits_{o_j \in {\Bbb N}(o_i)} {{CIS(o_i,o_j)}{Q_j}} } \right\|}^2}}
\end{split}
\end{equation}

In the objective function, the item coupling and users' rating preferences are integrated together. Specifically, the first part reflects the subjective rating preferences and the last part reflects the item coupling. This means the users' rating preferences and item coupling act jointly to make recommendations. In addition, another distinct advantage is that, when we do not have ample rating data, it is still possible to make satisfactory recommendations via leveraging the implicit coupling information.

We optimize the above objective function by minimizing $L$ through the gradient decent approach:
\begin{equation}\label{gradientDescentP}
\frac{{\partial L}}{{\partial {P_u}}} = \sum\limits_{o_i} {{I_{u,o_i}}} ({r_m} + {P_u}{Q_i}^T - {R_{u,o_i}}){Q_i} + \lambda {P_u}
\end{equation}

\begin{equation}\label{gradientDescentQ}
\begin{split}
& \frac{{\partial L}}{{\partial {Q_i}}} = \sum\limits_u {{I_{u,o_i}}} ({r_m} + {P_u}{Q_i}^T - {R_{u,o_i}}){P_u} + \lambda {Q_i}+\\
& \alpha ({Q_i} - \sum\limits_{o_j \in {\Bbb N}(o_i)} {{CIS(o_i,o_j)}{Q_j}} ) - \\& \alpha \sum\limits_{o_j:o_i \in {\Bbb N}(o_j)} {{CIS(o_j,o_i)}({Q_j} - \sum\limits_{o_k \in {\Bbb N}(o_j)} {{CIS(o_j,o_k)}{Q_k}} )}
\end{split}
\end{equation}
where $I_{u,o_i}$ is an logical function indicating whether the user has rated item $o_i$ or not. $CIS(o_i,o_j)$ is the coupled similarity of items $o_i$ and $o_j$. ${\Bbb N}(o_i)$ represent the item neighborhood.

The optimum matrices $P$ and $Q$ can be computed by the above gradient descent approach. Generally, the CIMF model starts by computing item coupling based on the objective content, then commences an iteration process for optimizing $P$ and $Q$ until convergence, according to Eqn. \ref{gradientDescentP} and \ref{gradientDescentQ}. Once $P$ and $Q$ are learned, the ratings for user-item pairs $(u,o_i)$ can be easily predicted by Eqn. \ref{model-basicMF}.

\section{Experiments and Results}
In this section, we evaluate our proposed model and compare it to the existing approaches respectively, using the MovieLens and Book-Crossing \cite{Ziegler:2005:bookcrossing} data sets.
\subsection{Data Sets}
The MovieLens data set has been widely explored in RS research in the last decade. The MovieLens 1M data set consists of 1,000,209 anonymous ratings of approximately 3,900 movies made by 6,040 MovieLens users who joined MovieLens in 2000. The ratings are made on a 5-star scale and each user has at least 20 ratings. The movies have titles provided by the IMDB (including year of release) and a special ``genre" attribute which is applied to compute the item couplings.

Similarly, collected by Cai-Nicolas Ziegler from the Book-Crossing community, the Book-Crossing data set involves 278,858 users with demographic information providing 1,149,780 ratings on 271,379 books. The ratings range from 1 to 10 and the books' ``book-author", ``year of publication" and ``publisher" have been used to form the item couplings.

\begin{table*}[htbp]
  \centering
  \caption{MF Comparison of MovieLens and Book-Crossing Data Sets}
    \begin{tabular}{|c|c|c|c|c|c||c|}
    \hline
    Data Set & Dim & Metrics & PMF (Improve) & ISMF (Improve)  & RSVD (Improve)  & CIMF \\ \hline
    \multirow{6}{*}{MovieLens} & \multirow{2}{*}{100D} & MAE & 1.1787(27.85\%) & 1.1125 (21.23\%) & 1.1076 (20.74\%) & \textbf{0.9002}\\ \cline{3-7}

    & & RMSE &1.7111 (70.53\%) & 1.5918 (58.60\%) & 1.5834 (57.76\%) & \textbf{1.0058}\\\cline{2-7}

    & \multirow{2}{*}{50D} & MAE &1.1852 (17.53\%) & 1.1188 (10.89\%) & 1.1088 (9.89\%) & \textbf{1.0099}\\\cline{3-7}

    & & RMSE &1.8051 (55.00\%) & 1.6103 (35.52\%) & 1.5835 (32.84\%) & \textbf{1.2551}\\ \cline{2-7}
	
    & \multirow{2}{*}{10D} & MAE &1.2129 (16.33\%) & 1.1651 (11.55\%) & 1.1098 (6.02\%) & \textbf{1.0496}\\ \cline{3-7}

    & & RMSE &1.8022 (48.35\%) & 1.7294 (41.07\%) & 1.5863 (26.76\%)& \textbf{1.3187}\\ \cline{2-7}
    \hline
     \hline

    \multirow{6}{*}{Book-Crossing} & \multirow{2}{*}{100D} & MAE &1.5127 (3.64\%) & 1.5102 (3.39\%) & 1.5131 (3.68\%)  & \textbf{1.4763}\\ \cline{3-7}
    & & RMSE &3.7455 (0.69\%) & 3.7397 (0.11\%)& 3.7646 (2.60\%) & \textbf{3.7386}\\ \cline{2-7}

    & \multirow{2}{*}{50D} & MAE &1.5128 (3.65\%) & 1.5100 (3.37\%) & 1.5131 (3.68\%) & \textbf{1.4763}\\\cline{3-7}
    & & RMSE &3.7452 (0.80\%) & 3.7415 (0.43\%) & 3.7648 (2.76\%) & \textbf{3.7372}\\ \cline{2-7}

    & \multirow{2}{*}{10D} & MAE &1.5135 (3.72\%) & 1.5107 (3.44\%) & 1.5134 (3.71\%) &  \textbf{1.4763}\\ \cline{3-7}
    & & RMSE &3.7483 (0.85\%) & 3.7440 (0.42\%) & 3.7659 (2.61\%) & \textbf{3.7398}\\\cline{2-7}
    \hline
    \end{tabular}
  \label{tab:comparisons_mf}
\end{table*}

\subsection{Experimental Settings}
The 5-fold cross validation is performed in our experiments. In each fold, we have 80\% of data as the training set and the remaining 20\% as the testing set. Here we use Root Mean Square Error (RMSE) and Mean Absolute Error (MAE) as evaluation metrics:

\begin{equation}
RMSE = \sqrt {\frac{{\mathop \sum \nolimits_{\left( {u,o_i} \right)|{R_{test}}} {{({r_{u,o_i}} - {{\hat r}_{u,o_i}})}^2}}}{{\left| {{R_{test}}} \right|}}} {\text{}}
\end{equation}

\begin{equation}
MAE = \frac{{\mathop \sum \nolimits_{\left( {u,o_i} \right)|{R_{test}}} \left| {{r_{u,o_i}} - {{\hat r}_{u,o_i}}} \right|}}{{\left| {{R_{test}}} \right|}}
\end{equation}
where $R_{test}$ is the set of all pairs $(u,o_i)$ in the test set.

To evaluate the performance of our proposed CIMF, we consider five baseline approaches based on a user-item rating matrix: (1) the basic probabilistic matrix factorization (PMF) approach \cite{SalMnih08_PMF}; (2) the singular value decomposition (RSVD) \cite{svd-gene} method; (3) the implicit social matrix factorization (ISMF) \cite{DBLP:implicit_social_SIGIR13} model which incorporates implicit social relationships between users and between items;(4) user-based CF (UBCF) \cite{SuK09_survey_CF}; and (5) item-based CF (IBCF) \cite{Deshpande:2004:ITN_recommendation}.

The above five baselines only consider users' rating preferences on items but ignore item coupling. In order to completely evaluate our method, we also compare the following three models PSMF, CSMF and JSMF, which respectively augment MF with the Pearson Correlation Coefficient, and the Cosine and Jaccard similarity measures to compute items' implicit relationships, based on the objective attribute information.

\subsection{Experimental Analysis}
We respectively evaluate the effectiveness of the proposed CIMF model by comparing it with the above methods.

\subsubsection{Superiority over MF Methods}
It is well known that MF methods are popular and successful in RS, hence, in this experiment, we compare our proposed CIMF model with the existing MF methods. In Table \ref{tab:comparisons_mf}, different latent dimensions regarding MAE and RMSE metrics on MovieLens are considered to evaluate the proposed CIMF model. In general, the experiments demonstrate that our proposed CIMF outperforms the other three MF baselines. Specifically, when the latent dimension is set to 10, 50 and 100, in terms of MAE, our proposed CIMF can reach improvements of 16.33\%, 17.53\% and 27.85\% compared with the PMF method. Regarding RMSE, the improvements reach up to 48.35\%, 55.00\% and 70.53\%. Similarly, CIMF can respectively improve by 6.02\%, 9.89\%, 20.74\% regarding MAE, and 26.76\%, 32.84\%, 57.76\% regarding RMSE over the RSVD approach. In addition to basic comparisons, we also compare our CIMF model with the latest research outcome ISMF which utilizes the implicit relationships between users and items based on the rating matrix by Pearson similarity. From the experimental result, we can see that CIMF can respectively improve by 11.55\%, 10.89\% and 21.23\% regarding MAE, and by 41.07\%, 35.52\%, 58.60\% regarding RMSE for different dimensions, 10, 50 and 100.

Similarly, we depict the effectiveness comparisons with respect to different methods on the Book-Crossing data set in Table \ref{tab:comparisons_mf}. We can clearly see that our proposed CIMF method outperforms all its counterparts in terms of MAE and RMSE. Specifically, when the latent dimension is set to 10, 50 and 100, in terms of MAE, our proposed CIMF can reach improvements of 3.72\%, 3.65\% and 3.64\% compared with the PMF method. Regarding RMSE, CIMF also improves slightly by 0.85\%, 0.80\% and 0.69\%. Similarly, CIMF can respectively improve by 3.71\%, 3.68\%, 3.68\% regarding MAE, and 2.61\%, 2.76\%, 2.60\% regarding RMSE over the RSVD approach. In addition, from the experimental results, compared with the latest research outcome ISMF, we can see that CIMF can respectively improve by 3.44\%, 3.37\%, 3.39\% regarding MAE, and by 0.42\%, 0.43\%, 0.11\% regarding RMSE for different dimensions, 10, 50 and 100.

Based on the experimental results on the MovieLens and Book-Crossing data sets, we can conclude that our CIMF method not only outperforms the traditional MF methods PMF and SVD, but also performs better than the state-of-the-art model ISMF in terms of MAE and RMSE metrics. Furthermore, the prominent improvements are the result of considering item couplings.

\begin{table*}[htbp]
\centering
  \caption{CF Comparison of MovieLens and Book-Crossing Data Sets}
  \label{tab:comparisons_cfs}
\begin{tabular}{|c|c|c|c||c|}\hline
Data Set & Metrics & UBCF (Improve)  & IBCF (Improve) & CIMF\\\hline
\multirow{2}{*}{MovieLens} & MAE & 0.9027 (0.25\%)& 0.9220 (2.18\%) & \textbf{0.9002}\\\cline{2-5}
& RMSE & 1.0022 (-0.36\%) & 1.1958 (19.00\%) & \textbf{1.0058}\\
\hline
\hline
\multirow{2}{*}{Book-Crossing} & MAE & 1.8064 (33.01\%) & 1.7865	(31.02\%) &\textbf{1.4763}\\\cline{2-5}
& RMSE & 3.9847	(24.61\%) &	3.9283 (18.97\%) & \textbf{3.7386}\\\hline
\end{tabular}
\end{table*}

\begin{figure*}
  \centering
  \subfigure[MAE on MovieLens]{
    \label{fig:subfig-strategy-mae-movielens} 
    \includegraphics[width=0.24\textwidth]{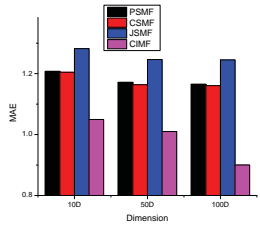}}
  \subfigure[RMSE on MovieLens]{
    \label{fig:subfig-strategy-rmse-movielens} 
    \includegraphics[width=0.24\textwidth]{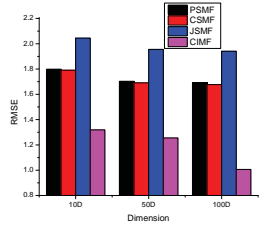}}
  \subfigure[MAE on Book-Crossing]{
    \label{fig:subfig-strategy-mae} 
    \includegraphics[width=0.24\textwidth]{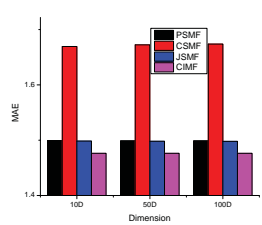}}
  \subfigure[RMSE on Book-Crossing]{
    \label{fig:subfig-strategy-rmse} 
    \includegraphics[width=0.24\textwidth]{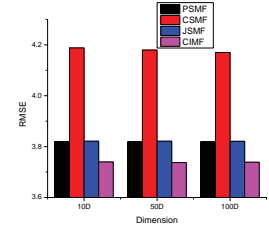}}
    \caption{Superiority over Hybrid Methods on MovieLens and Book-Crossing Data Sets}
     \label{fig-strategy-movielens}
\end{figure*}

\subsubsection{Superiority over CF Methods}
In addition to the MF methods, we also compare our proposed CIMF model with two different CF methods, UBCF and IBCF. In this experiment, we fix the latent dimension to 100 for our proposed CIMF model. On the MovieLens data set, the results in Table \ref{tab:comparisons_cfs} indicate that CIMF can respectively improve by 0.25\% and 2.18\% regarding MAE. Regarding RMSE, CIMF can improve by 19.00\% compared with IBCF, and slightly decreases by 0.36\% compared with UBCF but it is still comparable. Similarly, on the Book-Crossing data set, the results show that CIMF can respectively reach improvements of 33.01\%, 31.02\% regarding MAE, and 24.61\%, 18.97\% regarding RMSE compared with UBCF and IBCF. Therefore, this experiment clearly demonstrates that our proposed CIMF performs better than the traditional CF methods, UBCF and IBCF. The consideration of item couplings in RS contribute to these improvements.

\subsubsection{Superiority over Hybrid Methods}
In order to demonstrate the effectiveness of our proposed model, we compare it with three different hybrid methods, PSMF, CSMF and JSMF, which respectively augment MF with the Pearson Correlation Coefficient, and the Cosine and Jaccard similarity measures.

From the results shown in Fig. \ref{fig-strategy-movielens} on the MovieLens data set, generally we can clearly see that the coupled similarity method CIMF largely outperforms the three different comparisons with PSMF, CSMF and JSMF in terms of MAE and RMSE. Specifically, on the MovieLens data set for three different dimensions 10, 50 and 100, CIMF can respectively reach an improvement of 47.87\%, 44.72\% and 68.72\% regarding RMSE compared to PSMF. In terms of MAE, CIMF also can increase by 15.79\%, 16.14\% and 26.48\% compared to PSMF. Similarly, compared to CSMF on the MovieLens data set, CIMF can improve by 47.22\%, 43.57\% and 67.14\% regarding RMSE, while for MAE, the improvement can reach up to 15.49\%, 15.38\% and 26.03\%. Additionally, CIMF also performs better than JSMF, the respective improvements regarding RMSE being 74.18\%, 70.02\% and 93.47\%, while regarding MAE, CIMF also improves by 23.23\%, 23.63\% and 34.48\%.

 On the Book-Crossing data set, the results in Fig. \ref{fig-strategy-movielens} also indicate that the coupled similarity method CIMF constantly performs better than corresponding comparison methods regarding RMSE and MAE. Specifically, for three different dimensions 10, 50 and 100, CIMF can respectively reach an improvement of 7.91\%, 8.10\% and 8.01\% regarding RMSE compared to PSMF. In terms of MAE, CIMF also can slightly improve by 2.25\%, 2.21\% and 2.22\%, compared to PSMF. Similarly, compared to CSMF, on the MovieLens data set, CIMF can improve by 44.82\%, 44.24\% and 43.13\% regarding RMSE, while for MAE, the improvement can reach up to 19.31\%, 19.62\% and 19.75\%. Additionally, CIMF also performs better than JSMF, the respective improvements regarding RMSE being 8.15\%, 8.4\% and 8.22\%, while regarding MAE, CIMF also slightly improves by 2.22\%, 2.18\% and 2.16\%.

From these comparisons, we can conclude that our proposed CIMF model is more effective than the three different hybrid methods, PSMF, CSMF and JSMF.


\section{Conclusion}
The significant implicit information within items was studied for solving the \textit{cold start} and data \textit{sparsity} challenges in RS. To capture the implicit information, a new coupled similarity method based on items' subjective attribute spaces is proposed. The coupled similarity method discloses the traditional ``iid" assumption and deeply analyzes the intrinsic relationships within items. Furthermore, a coupled item-based matrix factorization model is proposed, which incorporates the implicit relations within items and the explicit rating information. The experiments conducted on the real data sets demonstrate the superiority of the proposed coupled similarity and CIMF model. Moreover, the experiments indicate that the implicit item relationships can be effectively applied in RS. Our work also provides in-depth analysis for the implicit relations within items and greatly extends the previous matrix factorization model for RS. Other aspects for enhancing our recommendation framework such as data characteristics or implicit relationships between users, will be investigated in the future.

\section{Acknowledgments}
We appreciate the valuable comments from all the reviewers.

\bibliography{ref}
\bibliographystyle{aaai}

\end{document}